\pdfoutput=1
\documentclass[11pt]{article}

\PassOptionsToPackage{table}{colortbl}  % use this to remove error.
\PassOptionsToPackage{table}{xcolor}  % use this to remove error.
\usepackage{amsmath}
\usepackage{amsfonts} 
\usepackage{enumitem}   
\usepackage{comment}
\usepackage{tabularx} % Required for the tabularx environment
\usepackage{booktabs} % For prettier tables
\usepackage{subcaption} % Required for sub-figures
\usepackage{tcolorbox} % for text boxes.
\usepackage{multirow}

\usepackage[normalem]{ulem}

\usepackage{stackengine}
\newcommand\barbelow[1]{\stackunder[1.2pt]{$#1$}{\rule{.8ex}{.075ex}}}

\usepackage[preprint]{acl}

\usepackage{times}
\usepackage{latexsym}

\usepackage[T1]{fontenc}
\usepackage[utf8]{inputenc}

\usepackage{microtype}

\usepackage{inconsolata}

\usepackage{graphicx}
\usepackage{subcaption}
\usepackage{array}
\usepackage{makecell}

\definecolor{orange_color}{HTML}{fbd0b9}
\definecolor{blue_color}{HTML}{a3cee3}
\newcommand{\positive}[1]{\cellcolor{orange_color!70}#1}
\newcommand{\negative}[1]{\cellcolor{blue_color!70}#1}

\title{How Personality Traits Influence Negotiation Outcomes?\\
A Simulation based on Large Language Models}

\author{Yin Jou Huang \and  Rafik Hadfi \\
  Graduate School of Informatics, Kyoto University, Kyoto, Japan\\
  \texttt{huang@nlp.ist.i.kyoto-u.ac.jp}, \texttt{rafik.hadfi@i.kyoto-u.ac.jp}\\}

\begin{document}

\maketitle

\begin{abstract}
Psychological evidence reveals the influence of personality traits on decision-making. For instance, agreeableness is generally associated with positive outcomes in negotiations, whereas neuroticism is often linked to less favorable outcomes. This paper introduces a simulation framework centered on large language model (LLM) agents endowed with synthesized personality traits. The agents negotiate within bargaining domains and possess customizable personalities and objectives. The experimental results show that the behavioral tendencies of LLM-based simulations can reproduce behavioral patterns observed in human negotiations. The contribution is twofold. First, we propose a simulation methodology that investigates the alignment between the linguistic and economic capabilities of LLM agents. Secondly, we offer empirical insights into the strategic impacts of Big Five personality traits on the outcomes of bilateral negotiations. We also provide an in-depth analysis based on simulated bargaining dialogues to reveal intriguing behaviors, including deceitful and compromising behaviors. 
\end{abstract}
\section{Introduction}

% LLM and decision making.
Large language models (LLMs) have demonstrated their capacity to emulate diverse human traits~\cite{park2022social, serapio2023personality}. Such models can simulate intricate behaviors and provide valuable insights into various aspects of human cognition. Decision-making is an example of cognitive processes that have long fascinated psychologists and economists. Economic theory posits that decisions assume a certain level of rationality and comprehension of available options~\cite{gibbons1992game}. However, behaviorists contend that humans are not entirely rational but are influenced by psychological factors~\cite{evans2014rationality}, cognitive biases~\cite{daniel2017thinking}, and personality traits~\cite{bayram2017decision}.

% negotiation and personality.
In this paper, we investigate the extent to which LLMs can simulate human decision-making across individuals with varying personality traits. We specifically focus on negotiations between two LLM agents with customizable personality traits. Evidence suggests that certain personality traits may give advantages in negotiation settings \cite{falcao2018big, barry1998bargainer, amanatullah2008negotiators}. For instance, agreeableness tends to result in a slight disadvantage in competitive negotiations while being an advantage in cooperative settings \cite{falcao2018big}. In the context of LLMs, we attempt to answer this long-standing question in psychology: \textit{``How do personality traits affect the outcomes of negotiations?"}

% Proposed Method.
%% Figure: system overview.
\begin{figure*}
    \centering
    \includegraphics[width=\textwidth]{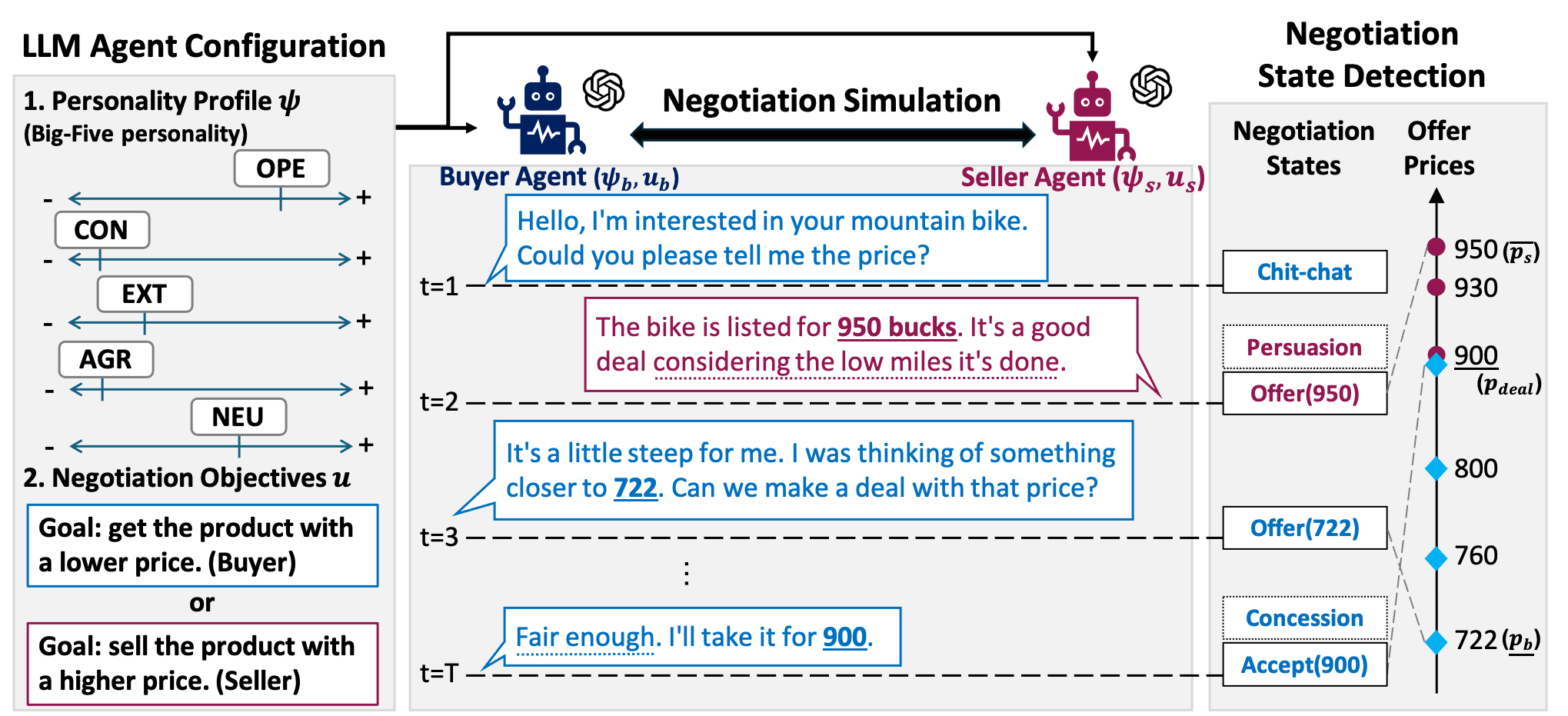}
    %\vspace{-7mm}
    \caption{Overview of the LLM-based negotiation simulation framework.}
    \vspace{-3mm}
    \label{fig:overview}
\end{figure*}

%% LLM negotiation agent.
To address this question, we propose a negotiation simulation framework incorporating LLM agents possessing synthetic personality traits and predefined negotiation objectives (Figure~\ref{fig:overview}).\footnote{The source code is available at \url{https://github.com/leslie071564/big5-llm-negotiator}.} For synthetic personalities, we use in-context learning to configure the agents with customizable personality profiles. Here, the profiles are based on Big Five personality theory \cite{costa1995domains, john1999big}.
For the negotiation objectives, we give LLM agents instructions that specify the negotiation task and goals. 
%% Negotiation simulation.
The LLM agents then engage in a negotiation by exchanging offers in the form of dialogue. 
% Scenario.
Specifically, we consider a competitive bargaining scenario between a buyer and a seller agent. In each round of the negotiation dialogue, we extract the offered price and strategy (if any) made in the dialogue utterances for evaluation and analysis.

% compare with human results.
By varying the personality traits of the agents, we observe changes in the behavioral patterns and outcomes of the negotiations.
We investigated which personality traits lead to better/worse outcomes. More importantly, we want to see whether the LLM-based simulation results align with the findings of research conducted on human subjects.
% findings.
Our experimental results show that the tendencies in LLM-based simulation generally align with behaviors observed in human experiments. This includes the effect of personality traits on negotiation outcomes and the relationship between personality traits and strategies. A case study based on synthesized bargaining dialogue reveals intriguing behavioral patterns such as deceiving behaviors, emotional appeal, and take-it-or-leave-it strategies. The results obtained in this work illustrate that LLMs can not only mimic styles of talking but are also capable of capturing human decision-making patterns.

% contribution.
The contribution of the paper is twofold. First, we propose a simulation framework that leverages LLM agents with linguistic and economic capabilities. Second, we offer insights into the effect of Big Five personality traits on simulated negotiation outcomes and compare these findings to empirical experiments from psychology.

\section{Related Work}

Recent advances in LLMs now allow the simulation of various human behaviors \cite {xie2024can,aher2023using,akata2023playing,serapio2023personality}. Decision-making is a particular type of behavior that is still challenging to reproduce using LLM agents because it relies on reasoning capabilities that they lack \cite{tamkin2021understanding}. Decisions generally entail choosing an action from various options in response to a particular situation, often reflecting personal preferences or beliefs \cite{simon1990bounded}. Moreover, real-world decisions are challenging because they are susceptible to environmental and cognitive constraints \cite{phillips2020decision}.

%Such factors primarily affect the quality of the decision-maker. Decisions and their quality can be assessed using various approaches. One could, for instance, rely on quantitative evaluations, specifically in economic terms, defined in terms of optimality \cite{koopmans1977concepts}. This presupposes that the decision maker is well aware of her choices, actions, preferences, and beliefs, though such awareness is not always guaranteed \cite{simon1990bounded, daniel2017thinking}. Alternatively, more qualitative approaches entail self-reported questionnaires \cite{urieta2021decision}. Examples of methods include the Proactive Decision-Making Questionnaire (PDMS) \cite{siebert2016developing}, the General Decision-Making Style (GDMS) \cite{scott1995decision}, and the Melbourne Decision-Making Questionnaire (MDMQ) \cite{mann1997melbourne}.

Since decision-making problems are wide, we choose to focus on negotiations as an example that we claim can be studied comprehensibly using LLMs. In a negotiation, two parties exchange bids and attempt to reach a mutual agreement \cite{raiffa1982art,jennings2001automated}. Looking at negotiations from the economics perspective, we often presuppose several assumptions, such as rationality \cite{evans2014rationality,ramansteer}. This assumption often fails when the negotiations are conducted through natural language, which conveys various aspects that cannot be studied economically, such as emotions or personality traits. There is evidence showing the effect of Big Five personality traits on decisions \cite{bayram2017decision, urieta2021decision, erjavec2019effect, toledo2023neurocircuitry, el2020personality}. Moreover, certain personality traits are considered disadvantageous in negotiations \cite{falcao2018big, amanatullah2008negotiators}. Extraversion and agreeableness, for example, constitute liabilities in competitive bargaining problems while being advantageous in cooperative settings \cite{barry1998bargainer}. Herein, we adopt an indirect economic analysis of negotiations to see whether personality instructions to LLMs really translate to genuine behavioral change in the underlying agents. We reproduce some of the known results of psychology in addition to results that are unique to LLMs.

Our analysis of the rationality of the LLM agents relies on economic metrics used in negotiation research \cite{baarslag2016learning}. Recent methods are tailored to LLMs and include the use of games and theory-of-mind (ToM) modeling \cite{ramansteer, davidson2024evaluating, qiu2024minddial}.

It is important to note that the simulation of human behavior using LLMs poses several risks, including the risk of data contamination \cite{oren2023proving}. For instance, if a dataset derived from a personality questionnaire is used to train a language model, the resulting personality scores may not accurately indicate whether the LLM's behavioral responses align with the personality traits specified in the prompt. To reduce the effect of such risks on our experimental design, we adopted canonical bargaining problems that are well-studied in negotiation research \cite{nash1950bargaining}. This choice renders the behavioral analysis of the LLM agents more tractable, particularly under the effect of personality traits. 

%-- Other findings explore the relationship between decision-making style, as measured by the Melbourne Decision Making Questionnaire, and personality traits, age, sex, and social position \cite{urieta2021decision}. The study investigates the association of decision-making styles with the Alternative Five Factor Personality Model (AFFPM), which includes domains such as Neuroticism, Extraversion, and Openness. In \cite{bayram2017decision}, the authors examine the relations between decision-making styles and personality traits among university students. In \cite{erjavec2019effect}, the authors explore the impact of personality traits and knowledge on the quality of decision-making. They found that individuals with lower levels of extraversion and agreeableness and higher levels of conscientiousness and openness tend to make better decisions. 
%Negotiation agents, with NLP -- not LLM
%\cite{rosenfeld2016negochat}. 
%\cite{kiruthika2020lifecycle}
%\cite{chawla2021towards}.
%\cite{toledo2023neurocircuitry,el2020personality}: survey  on negotiation %dialogue systems \cite{zhan2022let}

%---\im{(TODO-leslie: maybe add some paper that analyzes the behaviors of LLMs)}

\section{Methodology}

This section introduces our simulation framework based on LLM negotiation agents with synthetic personality traits. In Section \ref{sec:nego_model}, we formulate the negotiation model. In Section \ref{sec:agent_config}, we introduce the method to configure LLM negotiation agents by providing the instructions to set the personality traits and negotiation objectives. We then describe the process of simulating negotiation dialogues with the LLM agents in Section \ref{sec:dialog_creation}.

\subsection{Negotiation Model} \label{sec:nego_model}

We consider a classical \textit{bargaining} scenario in which a buyer and a seller negotiate over the price of a product. Typically, the buyer aims to reduce the purchase price while the seller seeks to maximize it, resulting in the competitive nature of the negotiation scenario. This is also an example of a \textit{zero-sum game} in which one party's gain leads to the other party's loss, showing the competitive nature of the task~\cite{gibbons1992game}.

In our LLM-based negotiation simulation scenario, the seller and the buyer are played by the LLM agents. Since our goal is to study the effects of personality traits on negotiation, we define the seller and buyer agents in terms of their psychological and economic profiles as in (Eq. \ref{nModel}).
\begin{equation}
    \begin{split}
    \text{Seller}\ s = (\psi_s, u_s) \\
    \text{Buyer}\ b = (\psi_b, u_b)
    \end{split}
    \label{nModel}
\end{equation}
% psychological profile.
The psychological profiles $\psi_s$ and $\psi_b$ will be instantiated with predefined personality traits following the Big Five model of personality. The Big Five model decomposes human personality into five dimensions: openness (\textbf{OPE}), conscientiousness (\textbf{CON}), extraversion (\textbf{EXT}), agreeableness (\textbf{AGR}), and neuroticism (\textbf{NEU}) \cite{costa1995domains,john1999big}. Each dimension is a spectrum with negative and positive polarities. The five dimensions encompass a comprehensive range of human personality patterns. 
% economic profile.
On the other hand, the economic profiles of the agents are reflected in their utility functions, denoted by $u_s$ for sellers and $u_b$ for buyers. A utility function is a mathematical way to describe the preferences or objectives of the agents depending on whether they are minimizing (buyer) or maximizing the price (seller) \cite{gibbons1992game}. 

The seller and the buyer negotiate in a dialogue $D$ around a product. The dialogue is a sequence of $T$ utterances $D=\{d_1, d_2, \ldots, d_T\}$. Each utterance $d_t$ is associated with a negotiation state $s_t$, the offer price $p_t$, and the negotiation strategy $\sigma_t$ adopted by the speaker at time $t$.

\subsection{LLM Agent Configuration} \label{sec:agent_config}

We configure an LLM agent with specific personality traits by introducing a personality instruction (Section \ref{sec:persona_intruction}) and a negotiation objective instruction (Section \ref{sec:nego_instruction}), with in-context learning.

\subsubsection{Personality Traits Instruction} \label{sec:persona_intruction}

% table: big5 dimension & positive negative & adjective.
\begin{table*}[t]
    \centering
    \resizebox{0.97\textwidth}{!}{%
    \begin{tabular}{ccc}
         \toprule
         \textbf{Dimension} & \textbf{Negative} & \textbf{Positive} \\ \midrule
         \textbf{OPE} & unimaginative, uncreative, unaesthetic, ... & imaginative, creative, aesthetic, ...\\ 
         \textbf{CON} & unsure, messy, irresponsible, ...& self-efficacious, orderly, responsible, ... \\
         \textbf{EXT} & unfriendly, introverted, silent, ... & friendly, extraverted, talkative, ... \\
         \textbf{AGR} & distrustful, immoral, stingy, ...& trustful, moral, generous, ... \\
         \textbf{NEU} & relaxed, at ease, easygoing, ... & tense, nervous, anxious, ... \\ 
         \bottomrule
    \end{tabular}
    }
    \vspace{-2mm}
    \caption{The Big Five personality dimensions and their corresponding personality-describing adjective pairs.}
    \vspace{-3mm}
    \label{tab:big-five}
\end{table*}

We assign synthetic Big Five personality profiles to LLM agents. That is, an agent $k$, with $k \in \{s, b\}$, possesses a $5$-dimensional personality profile $\psi_k$ defined as in (Eq. \ref{eq:personality_space}).
\begin{align}
    \label{eq:personality_space}
    \psi_k &= (\psi_k^{OPE}, \psi_k^{CON}, \psi_k^{EXT}, \psi_k^{AGR}, \psi_k^{NEU}) \nonumber \\ 
   \psi_k &\in \mathbb{L}^5 \\
    \mathbb{L} &= \{\text{-}, \text{+}\} \otimes \{\text{Low}, \text{Moderate}, \text{High}\} \nonumber
\end{align}
%-- old :
% \mathbb{L} &= \{-, +\} \times [1, 3]%
% \psi_k &\in \mathbb{L}^5  \\
% \mathbb{L} &= \{ ---, --, -, +, ++, +++ \} %\nonumber 

Each component of $\psi_k$ represents the polarity (negative or positive) and degree (low, moderate, or high) of each personality dimension~\cite{tian-etal-2018-polarity}. For instance, $\psi_k^{AGR}$ takes on one of the values in $\mathbb{L}$, which respectively represents a spectrum from highly disagreeable ($\text{-}\text{-}\text{-}$), moderately disagreeable ($\text{-}\text{-}$), lowly disagreeable ($\text{-}$), lowly agreeable (+), moderately agreeable (++), and highly agreeable (+++). Further, one can define a complete personality profile by sampling from the personality space $\mathbb{L}^5$ a vector $\psi_k$ such as (\textbf{OPE$\text{+}$, CON$\text{-}\text{-}\text{-}$, EXT$\text{-}$, AGR$\text{+}$, NEU$\text{+}\text{+}$}). 

Following previous work, we use personality-describing adjectives to set the personality traits \cite{serapio2023personality}. We use the list of $70$ bipolar adjective pairs proposed by \citet{goldberg1992development}, which are adjectives that statistically correlate with certain Big Five personality traits (Table \ref{tab:big-five}). For instance, prompting an LLM with adjectives such as \textit{unsure} and \textit{irresponsible} is likely to result in utterances with \textit{negative conscientiousness} traits. For each personality dimension in $\psi_k$, we randomly pick $n$ adjectives out of all the personality-describing adjectives associated with the polarity of the given dimension. Further, we apply the modifiers based on the degree of the personality traits. We use ``very'' as a modifier for a high degree and ``a bit'' for a low degree. No modifier is used for the moderate degree. Following this process, we use $5 \times n$ adjectives associated with any given personality profile $\psi_k$. 
We then generate a personality trait instruction with the template \textit{``You have following personality: $\$\{list\}$. Reflect your personality in the negotiation process.''}, where $list$ is a comma-separated list of the associated adjectives (including the modifiers). The personality traits instructions are given to the LLM agents through in-context learning~\cite{fu2023improving}.

\subsubsection{Negotiation Objective Instructions}
\label{sec:nego_instruction}

To configure the economic profiles of the LLM agents $u_s$ and $u_b$ in (Eq. \ref{nModel}), we incorporate negotiation objective instructions that define the negotiation goals of an agent. Specifically, we focus on the bargaining scenario where the seller agent aims to sell the product at a higher price, reaching its ideal price as closely as possible~\cite{raiffa1982art}. Conversely, the buyer agent seeks to secure a deal at a lower price and strives to achieve its ideal target price. The instructions are the following:
\vspace{-1mm}
\begin{itemize}
    \item \textbf{Buyer:} Act as a buyer and try to strike a deal for a $\$\{product\}$ with a lower price through conversation. You would like to pay for {$\barbelow{p}_b$}. Your reply should not be too long. 
    
    \item \textbf{Seller:} Act as a seller that sells a $\$\{product\}$, bargains with the buyer to get a higher deal price. Your reply should not be too long. Your listing price for this $\$\{product\}$ is {$\bar{p}_s$}. The detail of the product is the following: $\$\{description\}$.
\end{itemize}

Here, $product$ and $description$ are the product name and a short description of the product of the negotiations, and $\barbelow{p}_b$ and $\bar{p}_s$ are the ideal prices of the buyer and the seller, respectively. These linguistic instructions can theoretically be mapped into utility functions, which will later be used to evaluate negotiations. We avoid making assumptions about the shape of the utility functions, as the behaviors of the agents are primarily shaped by the LLM instructions, which may not follow any specific mathematical representation of their preferences.

\subsection{Negotiation Simulation} \label{sec:dialog_creation}

% Iterative/Alternative prompt feeding.
Using the methods in Section \ref{sec:agent_config}, we configure the buyer LLM agent and the seller LLM agent and conduct a negotiation simulation between them. The seller and buyer agents exchange offers alternatively, with the seller kick-starting the conversation with the fixed utterance \textit{``Hi, how can I help you?''}. After an utterance $d_t$ is generated, the response is fed to the other agent as a prompt to generate the next utterance $d_{t+1}$. The process continues until the termination condition is met. In this fashion, we collect a simulated negotiation dialogue $D = \{d_1, d_2, \ldots, d_T\}$. 

% state detector.
Following \citet{fu2023improving}, we introduce a dialogue state detector to extract negotiation-related information from each utterance. 
After the generation of each utterance $d_t$, the dialogue state detector takes $d_t$ and its context (previous $t-1$ utterances) as input and extracts the negotiation state $s_t$, current offer price $p_t$, and strategy $\sigma_t$ of $d_t$.
The negotiation state $s_t$ is one of the following:
\vspace{-2mm}
\begin{itemize}
    \item \textit{Offer}: the agent makes a price offer.
    \item \textit{Ponder}: the agent considers whether to accept or reject the other agent's offer.
    \item \textit{Accept}: the agent accepts the current offer.
    \item \textit{Deal-break}: the agent refuses the last offer or walks away from the negotiation.
      \item \textit{Chit-chat}: the agent engages in chatting that is not directly related to the negotiation, such as greetings.
\end{itemize}
In addition, we extract the current offer price $p_t$ (if any) and the strategy $\sigma_t$ of the current speaker in free text form. In this work, we use a third, separate LLM as the dialogue state detector. Details about the state detector can be found in Appendix \ref{PromptSettingsAp}.

% termination condition
For the termination condition, we terminate the negotiation dialogue if the negotiation state \textit{Accept} or a \textit{Deal-break} are reached. Also, we set a length limit of $T=T_{max}$ of the dialogue. If the length of the generated dialogue reaches this limit, we terminate the process and automatically regard it as a failed negotiation.

\section{Experiments}

In this section, we first provide details on the experimental settings (Section~\ref{sec:exp}) and the evaluation metrics (Section~\ref{EvalNegSection}).

\subsection{Experimental Settings}
\label{sec:exp}

%We introduce our experimental settings.
\paragraph{LLM Agents} For the buyer and seller agents, we adopt GPT-4 (gpt-4-0613) \cite{openai2023gpt4} as the choice of the LLM. We also conducted experiments based on Llama 3 and GPT-3.5, which can be found in Appendix \ref{LLMApendix}. The prompts of the agents can be found in Appendix \ref{PromptSettingsAp}.

\paragraph{Personality Instruction} For each agent $k$, we first generate a personality profile by randomly sampling from the personality space $\mathbb{L}^5$ (Eq. \ref{eq:personality_space}). For each Big Five dimension, we randomly select $n=3$ personality-describing adjectives associated with the sampled polarity and prepend the modifiers associated with the sampled degree to them.
The $5 \times n$ adjectives across all dimensions are then shuffled and given as instruction to agent $k$. Before using the LLM agents in negotiation simulations, we verify that the agents correctly reflect the given personality profiles with a personality test. See Appendix~\ref{sec:ipip} for details.

\paragraph{Negotiation Variables} We used the CraigsListBargain dataset~\cite{he-etal-2018-decoupling} to set several negotiation variables. It is a commonly used dataset of negotiation, consisting of bargaining dialogues in an online platform. We sampled a total of $60$ negotiation entries from the dataset. For each entry, we extract the name and the description of the product, the `listing price' of the seller, and the `target price' of the buyer. We use the listing price as the ideal price $\bar{p}_s$ for the seller and the target price as the ideal price $\barbelow{p}_b$ for the buyer. Note that in our setting, an agent's ideal price is not disclosed to the other party.

\paragraph{Dialogue Simulation} For the dialogue simulation process, we set a maximum length of $T_{max}=20$ rounds.
We use GPT-4 (gpt-4-0613) and the function calling module provided by OpenAI to implement the negotiation state detector. The full dialogue context is given to the negotiation state detection as input. The prompt for negotiation state detection can be found in Appendix \ref{PromptSettingsAp}. We collect a total of $1499$ negotiation dialogues with our simulation methodology.

% result table.
\begin{table*}[t]
%\small
\centering
\resizebox{\textwidth}{!}{%
\begin{tabular}{l|cc|cc|cc|cc|cc}
\toprule
\textbf{Personality} & \multicolumn{2}{c|}{\textbf{EXT}} & \multicolumn{2}{c|}{\textbf{AGR}} & \multicolumn{2}{c|}{\textbf{CON}} & \multicolumn{2}{c|}{\textbf{NEU}} & \multicolumn{2}{c}{\textbf{OPE}} \\ \hline
  & \textbf{Buyer} & \textbf{Seller} & \textbf{Buyer} & \textbf{Seller} & \textbf{Buyer} & \textbf{Seller} & \textbf{Buyer} & \textbf{Seller} & \textbf{Buyer} & \textbf{Seller} \\ 
\hline
Intrinsic Utility & 0.039 & -0.025 & \negative{\textbf{-0.056**}} & \negative{\textbf{-0.262**}} & \positive{0.044*} & \positive{\textbf{0.127**}} & 0.036 & \positive{\textbf{0.064**}} & 0.001 & -0.001 \\
Joint Utility & 0.001 & \positive{\textbf{0.069**}} & \positive{\textbf{0.054**}} & \positive{\textbf{0.118**}} & -0.016 & \positive{\textbf{0.072**}} & \negative{\textbf{-0.058**}} & 0.017 & 0.037 & \positive{\textbf{0.063**}} \\
Concession Rate & -0.009 & \positive{0.042*} & \positive{\textbf{0.051**}} & \positive{\textbf{0.261**}} & -0.010 & \negative{\textbf{-0.097**}} & -0.020 & \negative{\textbf{-0.074**}} & -0.007 & 0.016 \\
Success Rate & \positive{\textbf{0.072**}} & 0.036 & \positive{\textbf{0.065**}} & \positive{\textbf{0.052**}} & 0.026 & 0.001 & -0.007 & 0.021 & 0.008 & \positive{0.046*} \\
Negotiation Rounds & -0.024 & \negative{\textbf{-0.060**}} & \negative{\textbf{-0.136**}} & \negative{\textbf{-0.223**}} & \positive{0.083**} & 0.039 & 0.019 & -0.003 & 0.017 & -0.039 \\
\bottomrule
\end{tabular}}

\vspace{-1mm}
\caption{Spearman's correlation coefficients between negotiation metrics and Big Five personality traits. Asterisks indicate statistical significance, with * denoting \(p < 0.1\) and ** denoting \(p < 0.05\).}
\vspace{-1mm}
\label{table:newCorr}
\end{table*}

\subsection{Evaluation of the Negotiations}
\label{EvalNegSection}

% The other method:
%Use the ``final'' deal price from the actual data $p_{deal}$ as a measure of how ``good'' is a price, in general. That is, assume that $\barbelow{p}_s = \bar{p}_b = p_{deal}$ giving new formulations (\ref{UtilityFunctions2}).
%\begin{equation}
%\label{UtilityFunctions2}
%\begin{split}
%    u_s(p) = \frac{p - p_{deal}}{\bar{p}_s - p_{deal}} \\
%    u_b(p) = \frac{p_{deal} - p}{p_{deal} - \barbelow{p}_b} 
%\end{split}
%\end{equation}
%I believe this is close to your original formulation. It is easier since it uses what we already have and does not make assumptions about the agreement zone (i). \imp{Also, this utility function could be used to assess how good are the utility values of the actual experiments, assuming that the deal has already been determined in the human experiment.}

%\subsubsection {Evaluation Metrics}

We mainly evaluate the negotiations in terms of utility and whether the negotiations are successful or not \cite{lin2023opponent}. Recall that utility functions serve as mathematical tools for quantifying the quality of decision outcomes \cite{simon1990bounded}. In our study, we adopt economic metrics commonly used to evaluate negotiations \cite {baarslag2016learning}.

\paragraph{Intrinsic Utility} Based on the negotiation instructions, the utility of buyer and seller for a particular price $p$ is expressed in (Eq. \ref{UtilityFunctions}).
\begin{equation}
\label{UtilityFunctions}
\begin{split}
    u_s, u_b :  \mathbb{R}^{+} \rightarrow [0, 1] \\ 
    u_s(p) = \frac{p - \barbelow{p}_s}{\bar{p}_s - \barbelow{p}_s} \\
    u_b(p) = \frac{\bar{p}_b - p}{\bar{p}_b - \barbelow{p}_b} 
\end{split}
\end{equation}

As illustrated in the example of Figure \ref{fig:overview}, the prices $\barbelow{p}_s$ and $\bar{p}_s$ are the seller's reservation price and initial price, and $\barbelow{p}_b$ and $\bar{p}_b$ are the buyer's initial price and reservation price. Here, $\barbelow{p}_s$ is the price the seller is willing to accept without losing money. Similarly, the buyer's reservation price $\bar{p}_b$ is the maximum price it is willing to pay. Generally, the agreement zones of the agents are defined as the intersection between $[\barbelow{p}_b, \bar{p}_b]$ and $[\barbelow{p}_s, \bar{p}_s]$. We set $\barbelow{p}_s$ and $\bar{p}_b$ by assuming that the agreement zone is defined as $70\%$ of $[\barbelow{p}_b$, $\bar{p}_s]$.
%\footnote{Leslie: are we using this 100\%, 70\%,.. method?}. 
Second, $[\barbelow{p}_b, \bar{p}_b]$ and $[\barbelow{p}_s, \bar{p}_s]$ are private to the agents. Note that, in general, an offer is not guaranteed to fall within the intervals due to hallucinations \cite{ji2023survey}.

\paragraph{Joint Utility} We measure the fairness of negotiation outcomes using a joint utility function (Eq. \ref{JUEq}) inspired by Nash solution for bargaining problems \cite{luce1989games}. 
\begin{align}
    \label{JUEq}
	u_{sb}(p) = \frac{(p - \barbelow{p}_s) (\bar{p}_b - p)}{(\bar{p}_b - \barbelow{p}_s)^2} 
\end{align}
Such quantity is proportional to the product of the buyer's and seller's intrinsic utilities. Higher joint utility values indicate that the outcome is fair for both agents. For instance, an agreement price yielding a utility of $0.5$ for both agents is more equitable than an outcome yielding $0.8$ and $0.2$.

\paragraph{Concession Rate} Given the negotiation objectives, the offers could be assumed to undergo some form of decay akin to concessions. That is, an agent $k$ will make an offer at round $t \in [1, T]$ based on a discounted utility function (Eq. \ref{UtcEq}), with concession rate $c_k \in [0, 1]$. 
\begin{equation}
	\label{UtcEq}
	u_k^{(t)} = \barbelow{p}_k + ( \bar{p}_k - \barbelow{p}_k ) \times \bigg ( \frac{T - t}{T} \bigg)^{c_k}
\end{equation}
Applied to the utility functions of the buyer and seller (Eq. \ref{UtilityFunctions}), we obtain the concession rates (Eq. \ref{cRatesEq}).
\begin{equation}
\label{cRatesEq}
\begin{split}
 CR_s = \sum_{t=1}^T \log \bigg ( \frac{\bar{p}_s - p_t }{\bar{p}_s - \barbelow{p}_s} \bigg ) \\
  CR_b = \sum_{t=1}^T \log \bigg ( \frac{p_t - \barbelow{p}_b}{\bar{p}_b - \barbelow{p}_b} \bigg )
\end{split}
\end{equation}

%Without any assumptions about the shape of the discounted utility (\ref{UtcEq}), one could approximate the concession rate for both players using (\ref{cRatesEqNolog}).
%
%\begin{equation}
%\begin{split}
%\label{cRatesEqNolog}
%     CR_s = \frac{1}{nc_b} \sum_{t=1}^{nc_s}  \big ( \bar{p}_s - p_t  \big )\\
%     CR_b = \frac{1}{nc_s} \sum_{t=1}^{nc_s}  \big ( p_t  - \barbelow{p}_b \big )
%\end{split}
%\end{equation}

%Where $nc_b$ and $nc_s$ are the number of concessions per agent and could be set to the number of rounds.\footnote{Confirm difference between $T$ AND $nc_b$ and $nc_s$.}

%Possible interpretations of $CR_b$ and $CR_s$ will attempt to connect $\psi$ from \cite{tey2021impact} and ``Effects of the extremity of offers and concession rate on the bargaining outcomes.''.

%\imp{However, it seems that the interactions are time-independent since there is no mention of time in the instructions (?), or in other words, the interactions are short (what is the average length of rounds !?). If time has no influence, maybe use :
%\begin{align}
%CR_s = \frac{1}{T} \sum_{t=1}^T \frac{\bar{p}_s - p_s^{(t)}}{\bar{p}_s} 
%\end{align}}

Note that the estimation of concession rates from bid sequences depends on the round length and the linearity of the utility functions. The formulation in (Eq. \ref{cRatesEq}) represents one general form of concession estimation but could be refined to account for nonlinear utilities or dynamics bidding strategies \cite{baarslag2016learning}.

\paragraph{Negotiation Success Rate} The ratio (Eq. \ref{NSReq}) is that of successful negotiations $N_{succ}$ relative to the total number of negotiations $N$.
\begin{align}
    \label{NSReq}
	NSR = \frac{N_{succ}}{N}
\end{align}
\paragraph{Average Negotiation Round} The average (Eq. \ref{ANReq}) refers to the speed of successful negotiation \cite{lin2023opponent}.
\begin{align}
    \label{ANReq}
 ANR = \frac{1}{N_{succ}} \sum_{k=1}^{N_{succ}} T_k
\end{align}
 where $T_k$ is the number of rounds of the $k^{th}$ successful negotiation.

%\subsubsection{? Length of the utterances?}
%In \cite{he-etal-2018-decoupling}: ``Length is the number of utterances in a dialogue, thus encourages agents to chat as long as possible.'' It could be valuable in the interpretation of the results. Can we devise a metric that uses the length of the utterances as a function of the utility? For example, one could think of two negotiation problems $\mathcal{N}_1$ and $\mathcal{N}_2$ with the same personality traits and utility models, leading to the same deal but resulting from dialogues with different properties (length?).

\section{Results and Analysis}
In this section, we conduct an analysis of the negotiation outcomes based on the dialogues generated in the negotiation simulations.

\subsection{Negotiation Outcomes and Personality}
\label{CorrAnSection}

\begin{figure*}[t]
    \centering
    \includegraphics[width=\textwidth]{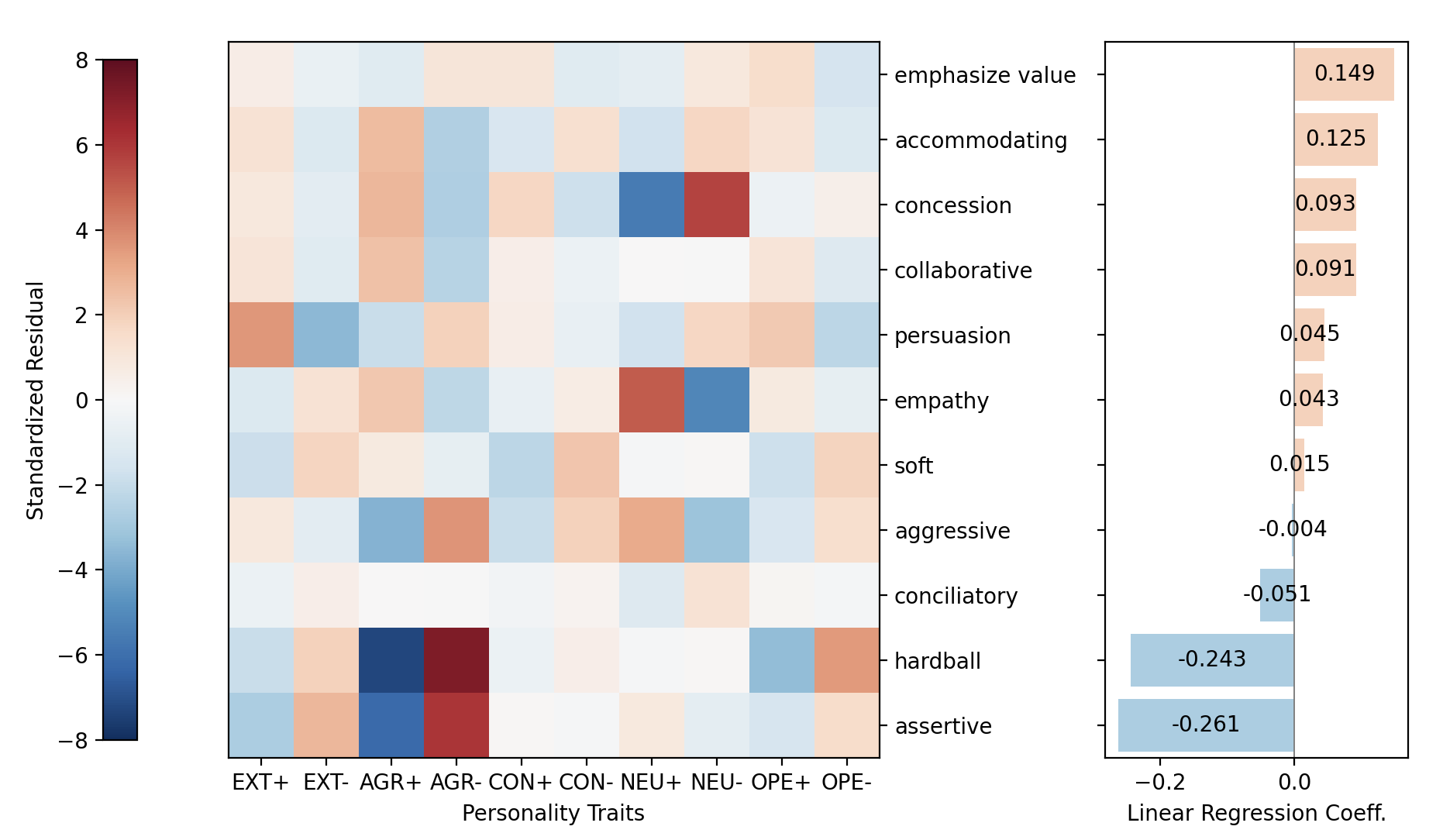}

    \vspace{-2mm}
    \caption{(Left) Positive and negative relationships between personality traits and strategies. Darker cells indicate higher statistical significance. (Right) Impact of strategies on the joint utility of the agreement price $p_{deal}$. Labels on the $y$ axes list the most common strategies among agents, occurring more than $20$ times within the entire negotiations.}
    \label{fig:Joint_Personality_Strategy}
    \vspace{-3mm}
\end{figure*}

First, we analyze the relationships between the negotiation outcomes and the personality traits of the agents. Table \ref{table:newCorr} illustrates Spearman's rank correlation coefficients between the economic metrics introduced in Section \ref{EvalNegSection} and Big Five personality traits.\footnote{The absolute value of the observed coefficients with statistical significance are all smaller than $0.3$, indicating only a mild correlation.
%The observed coefficients on Table \ref{table:newCorr}, while statistically significant, are generally weak, indicating minimal relationships between the personality traits and negotiation outcomes.%\footnote{\ed{It might be good to state this. And maybe the reasons: data? training?}}%https://www.researchgate.net/publication/267339455_Pilot_Validation_of_the_Tuberous_Sclerosis-Associated_Neuropsychiatric_Disorders_TAND_Checklist/figures?lo=1
}

% EXT.
Along the extraversion (\textbf{EXT}) dimension, we observed that extroverted buyer agents tend to result in more successful negotiations ($\rho=0.072^{**}$). On the other hand, an extroverted seller agent tends to result in a slightly higher joint utility, which indicates a fairer outcome for both negotiators.
% AGR.
Among the five dimensions, agreeableness (\textbf{AGR}) has the strongest impact on negotiation outcomes, especially for the seller. We find that being agreeable leads to a disadvantage in bargaining in terms of intrinsic utility ($\rho=-0.262^{**}$). 
%The negative correlation of extraversion (EXT) and agreeableness (AGR) with utility reflects a well-known effect of these traits in competitive settings, where they are considered liabilities in distributive bargaining encounters \cite{barry1998bargainer}.
However, agreeableness correlated positively with joint utility ($\rho=0.118^{**}$), concession rates ($\rho = 0.261^{**}$), and negotiation success ($\rho=0.052^{**}$), pointing to a propensity for cooperative behavior that benefits both negotiating parties. 
Also, agents tend to reach an agreement with fewer number of negotiation rounds ($\rho=-0.223^{**}$) with mutually satisfying outcomes. This aligns with the positive effect of agreeableness on the negotiators' distributive outcomes reported in \cite{sass2015personality}.
%For buyers, agreeableness (AGR) exhibited a negative correlation with the number of rounds or dialogue ($\rho=-0.136^{**}$), indicating that agreeable buyers engage in shorter negotiations, most likely ending in an agreement. On the other hand, agreeableness tends to correlate positively with concessions ($\rho=0.051^{**}$) and success of negotiation  ($\rho=0.065^{**}$), suggesting that more agreeable buyers are likely to make more concessions and reach agreements. 
%For sellers, agreeableness (AGR) correlated positively with joint utility ($\rho=0.118^{**}$), concession rates ($\rho = 0.261^{**}$), and negotiation success ($\rho=0.052^{**}$), pointing to a propensity for cooperative behavior that benefits both negotiating parties. 
%
% CON.
Aligning with previous findings \cite{barry1998bargainer}, we observed a positive correlation between conscientiousness (\textbf{CON}) and negotiation performance, especially for the seller (intrinsic utility, $\rho=0.127^{**}$).
%“Conscientiousness could be expected to have an important role since negotiation performance has often been associated with the preparation and the structured mindset.”
Conscientiousness is also manifested in the negotiation style of not being willing to concede ($\rho=-0.097^{**}$) on the seller side and a lengthier negotiation on the buyer side ($\rho=0.083^{**}$).
%In contrast, high conscientiousness (CON) among sellers is negatively correlated with concession rates ($\rho=-0.097^{**}$), which could imply that more conscientious sellers might concede less in order to obtain higher intrinsic gain ($\rho=0.127^{**}$). 

% NEU.
In the psychology literature, neuroticism (\textbf{NEU}) and openness (\textbf{OPE}) are expected to play a less predominant role in negotiations \cite{falcao2018big}. Similarly, we only observe some very weak correlation (with magnitude $<0.1$) along these dimensions. For instance, neurotic sellers tend to make fewer concessions ($\rho=-0.074^{**}$) and have a slightly higher utility gain ($\rho=0.064^{**}$), while open-minded sellers lead to enhanced joint gain ($\rho=0.063^{**}$). 
% conclusion.
These results point to how personality traits could impact the outcomes of negotiations with different effects observed based on the role of the negotiator.

\subsection{Analysis of Negotiation Strategies}

\begin{figure*}[ht]
    \centering
    \begin{subfigure}[b]{2\columnwidth}
        \includegraphics[width=\columnwidth]{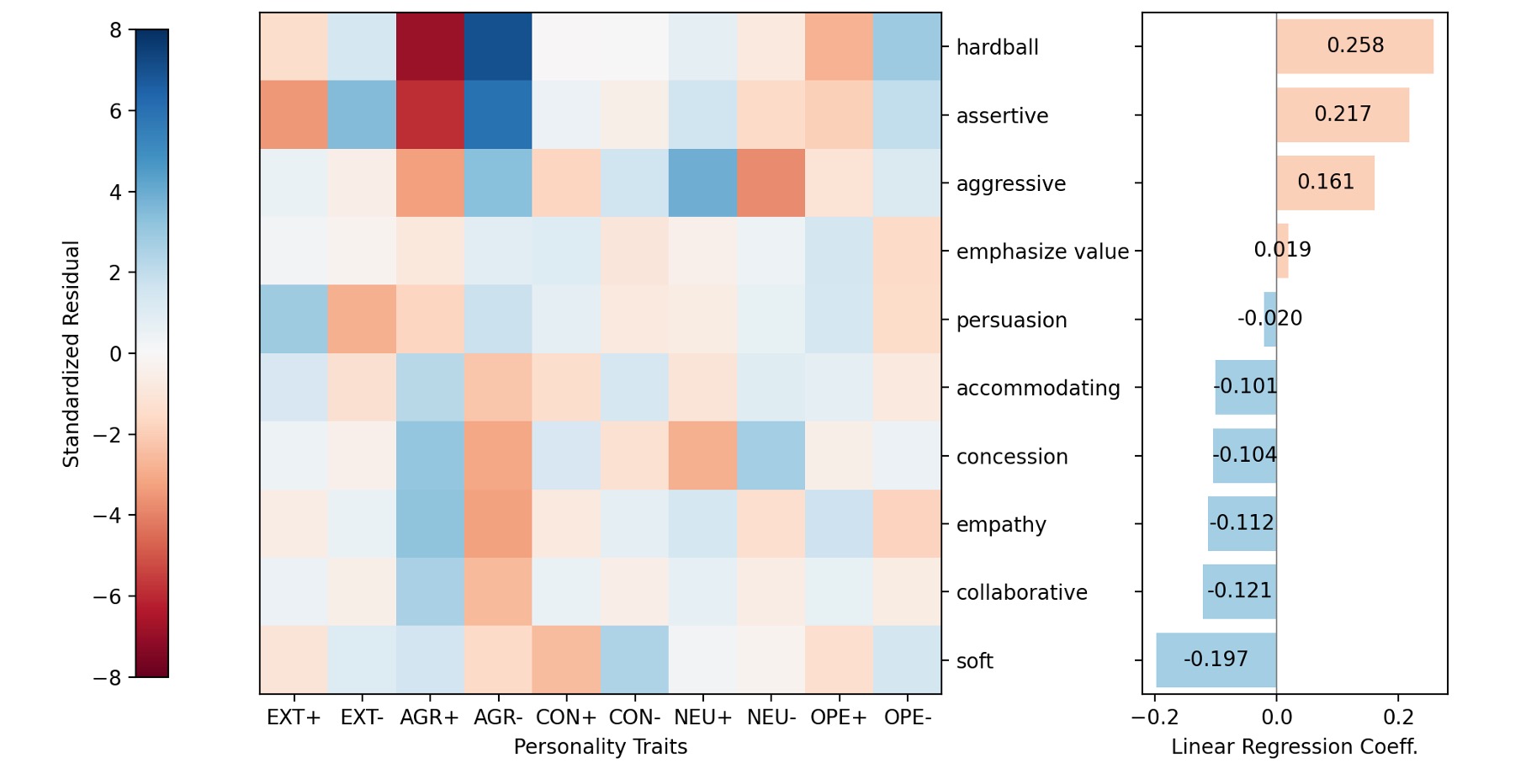}
        \vspace{-9mm}
        \caption{Seller}
        \label{fig:iu_strat2}
    \end{subfigure}
    %\hfill
    \vspace{5mm}
    \begin{subfigure}[b]{2\columnwidth} 
        \includegraphics[width=\columnwidth]{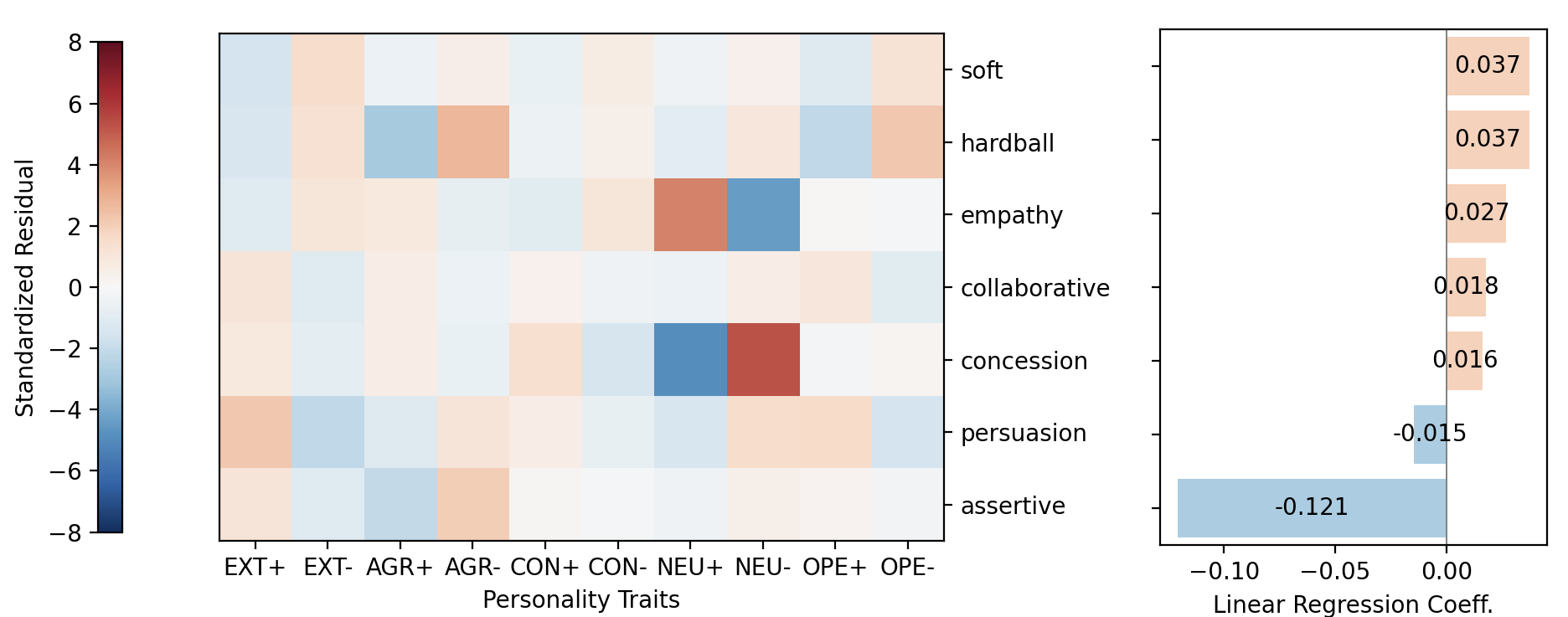}
        \caption{Buyer}
        \label{fig:iu_strat1}
    \end{subfigure}
    \vspace{-5mm}
    \caption{(Left) Relationship between personality traits and strategies. Darker cells indicate higher statistical significance. (Right) Impact of strategies on intrinsic utility gains. Labels on the $y$ axes list the most common strategies, occurring more than $20$ times within the entire negotiations.}
    \label{fig:utility_strat_visualize}
\end{figure*}

So far, the buyer and the seller are defined on the basis of personality traits and economic preferences. We now look at their strategies and the consequent effects on the joint negotiation outcomes. This investigation could answer questions of the type ``\textit{What are the optimal strategies for a neurotic buyer and a disagreeable, conscientious seller?}''. 

At each round, each agent adopts a specific strategy $\sigma_t$ that results in some price $p_t$. Given that the outcome of a successful negotiation is the final price $p_{deal}$, we are interested in the relationship between the adopted strategies and the joint utility of the agreement price, $u_{sb}(p_{deal})$. 
We manually inspected the strategies that occurred more than $20$ times within the entire negotiations and constructed the sets of strategies of the agents. We then conducted a linear regression between the strategy categories and the resulting joint utility values as illustrated in Figure \ref{fig:Joint_Personality_Strategy}.

On the right side of the Figure, collaborative strategies exhibit positive effects on the joint gain of both negotiators. For instance, accommodating ($\beta=0.125$) and conceding ($\beta=0.093$) strategies contribute positively to joint utility. Conversely, assertive and aggressive strategies ($\beta=-0.261$) negatively impact joint utility. This suggests that collaborative strategies are more beneficial for achieving mutual gains, while assertiveness diminishes them. The left side of Figure \ref{fig:Joint_Personality_Strategy} illustrates how these strategies depend on certain personality traits. We found that assertive strategies are more commonly expected from disagreeable agents (\textbf{AGR-}). On the other hand, neurotic traits (\textbf{NEU+}) do not lead to concessions, whereas a lack of neurotic traits (\textbf{NEU-}) leads to concessions. Finally, agents with neurotic traits seem to employ more empathetic strategies, while by contrast, agents with fewer neurotic traits will refrain from them. %The complete strategic evaluations of the intrinsic utilities of the buyer and seller can be found in Appendix \ref{NegoStrategyIndivAp}.

% below moved from the appendix.
%\section{Negotiation strategies of buyer and seller} \label{NegoStrategyIndivAp}
%In addition to the joint case illustrated in Figure~\ref{fig:Joint_Personality_Strategy}, 
In addition to joint utility, we have also looked at the individual gains of the buyer and the seller using their intrinsic utility functions $u_b(p_{deal})$ and $u_s(p_{deal})$ given the deal price $p_{deal}$.
We start with the strategy sets of the agents $k \in \{s, b\}$, namely $\Sigma_s$ and $\Sigma_b$. We found that $\Sigma_b \subseteq \Sigma_s$, indicating that sellers have additional strategies such as attempting to \textit{emphasize} the target price. Generally, at time $t$, agent $k$ adopts a strategy $\sigma_{k,t} \in \Sigma_k$, resulting in some price $p_t$, inferred from state $s_t$. Given that the outcome of a successful negotiation is the final price $p_{deal}$, we are interested in the relationship between the strategies of the agents $\Sigma_s$ and $\Sigma_b$ and the utility values of the deal price of the buyer $u_b(p_{deal})$ and seller $u_s(p_{deal})$. We conducted a regression between the strategies and the intrinsic utilities of the buyer and seller, illustrated in Figure \ref{fig:utility_strat_visualize}, aggregated across products and personality traits. Labels on the $y$ axes are the predominant, unified strategies of the players.

Figure \ref{fig:iu_strat2} (right) illustrates a nuanced impact of strategies on the seller's utility gain, highlighting a spectrum of strategies from cooperative to non-cooperative tactics (`soft' to `hardball'). Strategies positioned at the assertive end of the spectrum, such as assertiveness ($0.23$), exhibit a positive effect leading to an enhanced utility gain $u_s(p_{deal})$. On the cooperative end, strategies such as concession ($-0.11$) or empathy ($-0.13$) show negative impacts on the seller's utility. These results suggest that assertive strategies are more beneficial for sellers, while cooperative strategies tend to diminish the sellers' utility gains.
% Our test is that of independence:
% https://docs.scipy.org/doc/scipy-1.7.0/reference/reference/generated/scipy.stats.chi2_contingency.html
% https://www.scribbr.com/statistics/chi-square-tests/
We now look at the relationship between the strategies of the agents and their personality traits using Pearson’s Chi-square test. The heat-map cells in Figure \ref{fig:iu_strat2} (left) illustrate the positive (red) and negative (blue) strength of dependence between the traits ($x$ axis) and strategies ($y$ axis). For example, disagreeableness (\textbf{AGR-}) aligns with non-collaborative strategies, while neuroticism (\textbf{NEU}) aligns with aggressiveness.

For the buyer (Figure \ref{fig:iu_strat1} (right)), collaborative strategies such as softness ($0.06$) and empathy ($0.03$) exhibit positive effects on deal utility $u_b(p_{deal})$, indicating that strategies substantially enhance the buyer's gains. Concession shows moderate negative impacts ($-0.01$). Assertiveness ($-0.11$) is associated with the most significant reduction in the utility. Collaborative strategies also seem to be more beneficial for buyers, while assertiveness tends to diminish their utility gain. Figure \ref{fig:iu_strat1} (left) shows that a neurotic buyer (\textbf{NEU+}) tends to concede less and use empathy, while a less neurotic buyer (\textbf{NEU-}) will concede more and use less emphatic strategies.

From Figures \ref{fig:iu_strat1} and \ref{fig:iu_strat2}, it is possible to recover the strategic asymmetry that is characteristic of our bargaining game. Some strategies have opposing effects on the intrinsic utilities of the agents. For instance, assertiveness has a coefficient of $-0.1$ for the buyer and $0.23$ for the seller. Assertiveness, however, is detrimental to the joint utility as illustrated by the coefficient $-0.45$ in Figure \ref{fig:Joint_Personality_Strategy}.

\subsection{Analysis of Negotiation Length}

\begin{figure*}[t]
    \centering
    
    \begin{subfigure}[b]{0.49\textwidth}
        \centering
        \includegraphics[width=\textwidth]{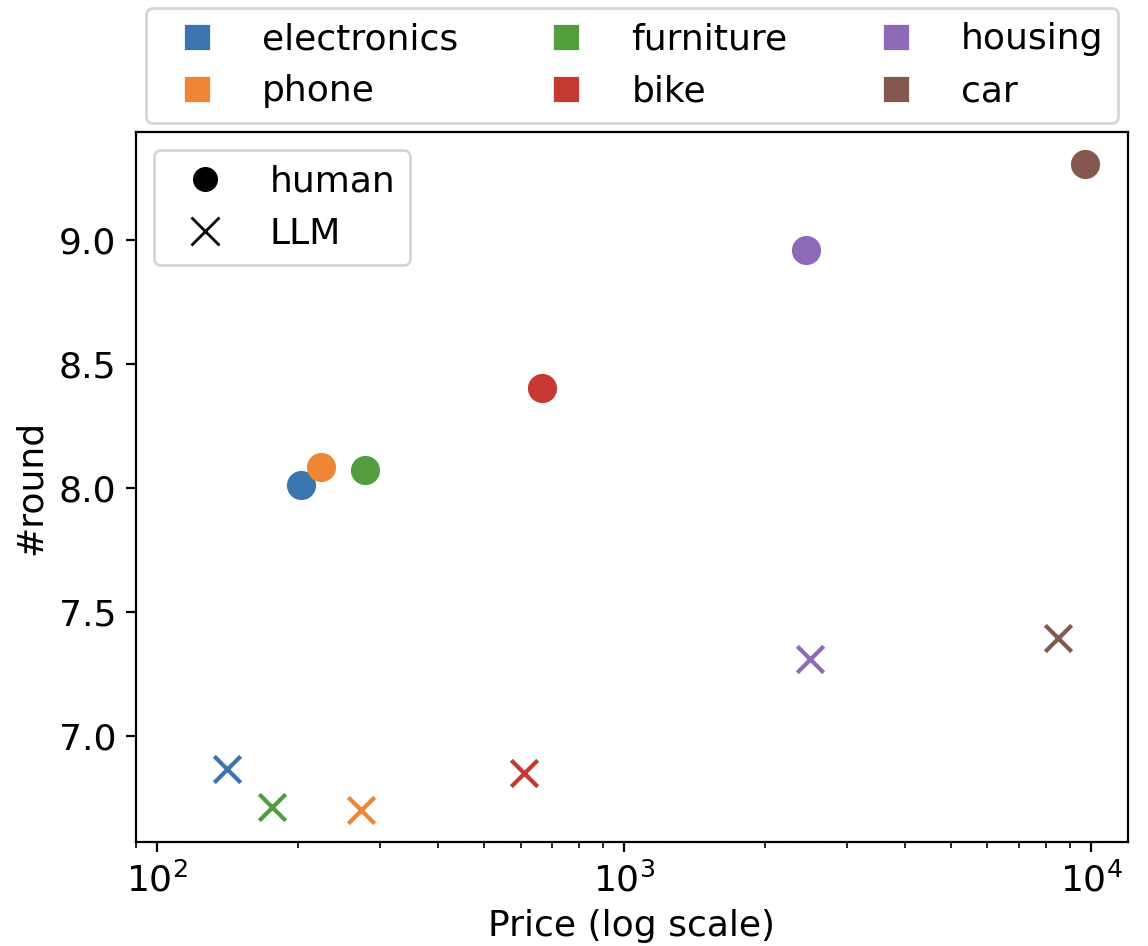}
        %\caption{Subfigure 1}
    \end{subfigure}
    \hfill
    \begin{subfigure}[b]{0.49\textwidth}
        \centering
        \includegraphics[width=\textwidth]{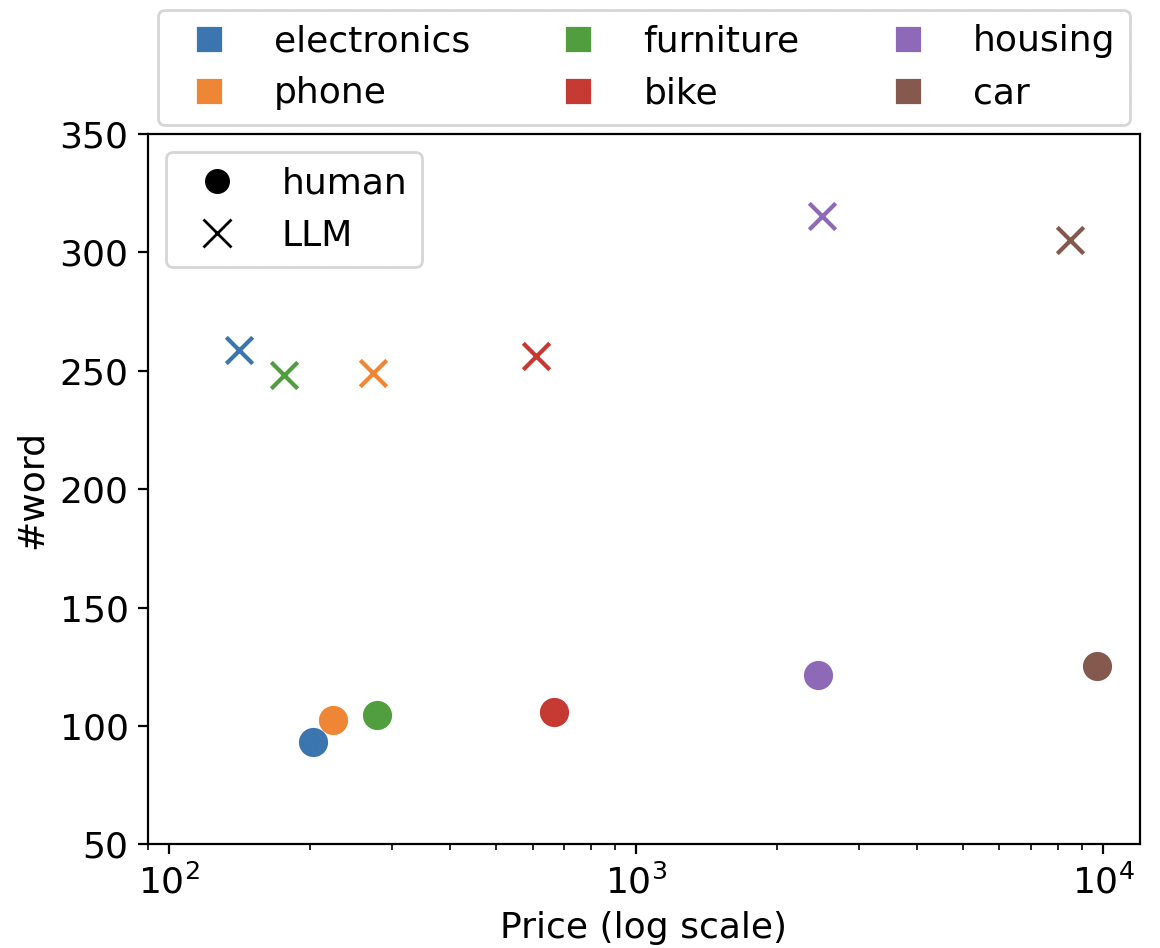}
        %\caption{Subfigure 2}
    \end{subfigure}

     \vspace{-2mm}
     \caption{The relations between the product prices and the number of negotiation rounds (Left) and the number of words (Right). The product prices are in log scale.}
     \vspace{-2mm}
    \label{fig:stats_product_category}
\end{figure*}

In this section, we conduct an analysis of the length of the negotiation dialogues.
First, we compare the average length of real-world negotiation dialogues in the CraigsListBargain dataset and the simulated dialogues generated by LLM negotiation agents. 
The real dialogues have an average length of $8.47$ rounds, while the LLM simulated negotiation dialogues have an average of $7.07$ rounds. 
In terms of the number of rounds, the simulated dialogues is $1.4$ rounds shorter than the real one on average.
We observe that LLM agents often delve into the bargaining task directly, while human negotiators often spend time greeting, showing appreciation, and chatting, which could cause a difference in the average negotiation rounds.
On the other hand, the real dialogues contain an average of $108.82$ words, and the simulated dialogues have an average of $272.17$ words.
Despite including ``Your reply should not be too long'' in the instruction (Section~\ref{sec:nego_instruction}), the LLM agents' utterances still tend to be longer than human utterances.

% table: put here to adjust the location of the table.
\begin{table*}[ht]
    \centering
    \resizebox{\textwidth}{!}{%
    \begin{tabularx}{\textwidth}{>{\hsize=1.1cm}X>{\hsize=1.1cm}X|>{\hsize=9.4cm}X|ccc}
    \toprule
    \textbf{Seller} &\textbf{Buyer} & \textbf{Negotiation Dialogue} & \textbf{\(\bar{p}_s\)} & \textbf{\(\barbelow{p}_b\)} & \textbf{\(p_{deal}\)} \\
    \midrule
    % removed deceiving example.
    %OPE+++\newline CON\text{-}\newline EXT\text{-}\text{-}\text{-}\newline 
    %AGR\text{-}\text{-}\text{-} \newline NEU\text{-}\text{-} \newline &
    %OPE\text{-}\text{-}\text{-}\newline CON+++\newline EXT+\newline AGR+\newline NEU\text{-}& 
    % \small{
    %    Buyer: Hello, I'm interested in your Home Security System. Could you please tell me the price?"\newline
    %    Seller: Hey. The price is \$165. But, I mean, you know how it is... \textbf{with something as valuable as your home security... would you look at offering more? Like, \$185 or so?} \newline
    %    ...}
    %    & 165 & 125 & 140 \\
    OPE+\newline CON\text{-}\newline EXT\text{-}\text{-}\text{-}\newline 
    AGR\text{-}\text{-}\text{-} \newline NEU+ \newline &
    OPE\text{-}\text{-}\text{-}\newline CON\text{-}\text{-}\text{-}\newline EXT\text{-}\text{-}\newline AGR\text{-}\newline NEU+& 
     \small{
        Buyer: Hello, I'm interested in your Stereo Speaker. Could you please tell me the price?\newline
        Seller: Yeah, the beautiful survivor from the Magnavox era? \textbf{Its price is \$80.} Solid piece, you won't find another one like this with such a unique gold tone finish, trust me. \newline
        ...\newline
        Seller: You are killing me here, really. I'm practically giving it away at this rate. But alright, let's meet halfway. How about \$60?\newline
        Buyer: ...Um... okay, it's a deal.
        }
        & 50 & 30 & 60 \\
    \midrule
    OPE\text{-}\text{-}\newline CON\text{-}\text{-}\text{-}\newline EXT+\newline 
    AGR\text{-}\text{-}\text{-} \newline NEU+ \newline &
    OPE\text{-}\text{-}\newline CON+\newline EXT\text{-}\text{-}\text{-}\newline AGR+\newline NEU\text{-}\text{-}\text{-}&
     \small{
        Seller: The iPhone 5S I'm selling is listed for a flat \$160. Can't really go any lower than that. Trust me, the phone is in like-new condition... 
        \newline 
        Buyer: Oh, I understand. I'm on a very tight budget at the moment, you see, and I was hoping to spend around \$144... \textbf{Also, it's not for me, it's for my elderly parent.} They aren't tech-savvy so even older models work for them...
        \newline
        Seller: Oh sure, pull at my heartstrings with the elderly parent bit. You know, some might fall for that. But let me tell ya, pal, life ain't a fairytale. \textbf{The iPhone's \$160, like I said earlier.} \newline
        Buyer: I understand and respect your stance. It's just my financial condition that's restricting... What about \$150, as my final offer? \newline
        Seller: \textbf{Look, this isn't a charity. It's \$160 or nothing.} ...\newline
    ...}
    & 160 & 144 & 160 \\
    \bottomrule
    \end{tabularx}
    }
    %\vspace{-2mm}
    \caption{Case analysis. The bold parts of the dialogue show deceptive and hardheaded negotiation behaviors.}
    \vspace{-4mm}
    \label{tab:case_study}
\end{table*}

Second, we look at the influence of product price on the negotiation dialogue length. The negotiation product in the CraigsListBargain dataset can be divided into six product types: electronics, phones, furniture, bikes, housing, and cars. Products of different types tend to have different prices. For instance, electronics and phones tend to cost around several hundred dollars, while cars are typically listed for up to ten thousand dollars. Figure~\ref{fig:stats_product_category} shows the average product price and negotiation length of each product type. For both real-world (\textbf{human}) and LLM simulated negotiations (\textbf{LLM}), we found a positive correlation between product price and negotiation lengths. 
We found a Pearson correlation coefficient of $0.194$
between product price and the number of negotiation rounds (\textbf{\#round}), and $0.242$ between product price and the number of words in the dialogue (\textbf{\#word}).
This indicates that the negotiation tends to be longer when the product price is higher. There is indeed evidence that human negotiators behave differently under varying stakes and constraints. Negotiators often adopt more deliberate and careful negotiation strategies as the stakes increase, which typically extend the negotiation process \cite{neale1991cognition}. This behavior is linked to the increased cognitive load and the desire to avoid unfavorable outcomes, leading to longer decision times. The stake in our experiments is defined as the price of the product under negotiation. We can hypothesize that increasing the number of items or adding complexity to the constraints or objectives will result in longer dialogues. 
%We also noted that for real-world negotiation, the Pearson correlation coefficient between product price and the number of negotiation rounds is $0.146$. On the other hand, the simulated negotiation has a Pearson correlation coefficient of $0.194$ between product price and number of negotiation rounds, which shows a slightly stronger positive correlation.

\subsection{Case Study}%: deception and intransigence}

We conducted a qualitative analysis of the negotiation results. Table \ref{tab:case_study} shows two distinct examples of dialogues generated in the negotiation simulations. 

% Example 1: deceiving behavior.
In the example on the first row, the seller and the buyer are negotiating over a stereo speaker. %, with the target price of $\$50$ and $\$30$, respectively. 
When the buyer asks about the price, the seller provides a price ($\$80$) that is higher than the listing price ($\$50$) and ends up striking a deal at $\$60$, which is higher than the listing price. We observe this kind of deceiving behavior in the negotiations involving disagreeable agents.
%The first example is a negotiation dialogue between a highly disagreeable (\text{AGR$\text{-}\text{-}\text{-}$}) seller and a buyer. We can see that when the buyer asks about the product's price, the seller answers a price ($\$180$) that is higher than the listing price ($\$160$) in the instruction. We discover the kind of deceiving behavior in the negotiations involving disagreeable agents.
%Example 2: Emotional appeal of buyer and hardball seller.
The second example is a negotiation on a second-hand iPhone. We can see that the buyer tries to emotionally appeal to the seller by mentioning her `my elderly parent,' but the seller is not persuaded and adopts a `take-it-or-leave-it' strategy. Despite the constant effort of the buyer to find common ground with the seller, the latter is not willing to change the offer price ($\$160$). We observe more similar `hardball' negotiation strategies with disagreeable and close-minded agents. At the end of the negotiation, the buyer concedes and accepts the seller's offer.
%The second example is a negotiation involving a non-neurotic Buyer (\textf{NEU$\text{-}\text{-}$}). We can observe a compromising behavior here: when the seller offers the price of $\$250$, the buyer concedes and accepts the deal easily (\textit{``And an extra $\$5$ won’t hurt, I guess.''}). On the other hand, in the third example of negotiation involving a seller of negative openness (\textf{OPE$\text{-}\text{-}\text{-}$}), we can see that the seller adopts a `take-it-or-leave-it' strategy. Despite the constant effort of the buyer to make a concession to find a common ground with the seller, the seller is not willing to change the offer price ($\$190$). This also results in a breakdown in negotiation, with the buyer leaving the negotiation table without reaching an agreement. 
%
The above examples showcase a range of negotiation behaviors such as deception, emotional appeal, and hard-headed behavior~\cite{baarslag2016learning}. This illustrates how specific personality traits influence negotiation styles.

\section{Conclusion}

This study introduced a novel simulation framework that integrates LLM negotiation agents with synthesized Big Five personality traits. These agents were deployed in traditional bargaining scenarios to simulate negotiation dialogues. Our experimental findings indicate that the behavioral patterns demonstrated by the LLM agents closely mirror those exhibited by human negotiators. Furthermore, a detailed analysis of synthesized bargaining dialogues highlighted distinct strategies and behaviors, including deceit, compromise, and stake assessment. These analyses augment our understanding of linguistic and economic agent interactions. 

Our contributions are twofold. Firstly, we established a robust simulation methodology to examine the linguistic and economic dynamics of LLM agents. Secondly, we provided comprehensive empirical evidence on the strategic impact of Big Five personality traits on negotiation outcomes, offering valuable insights into the mechanics of artificial negotiation agents.
The findings of this work have potential applications such as personalized assistants, chatbots, social dynamics simulations, and so on.
%In addition to potential expansions of the proposed economic and personality models, the present framework poses certain risks that require further investigation. One area of concern is the application of these LLM agents within assistive technologies, such as chatbots deployed on financial platforms.

\section*{Limitations}

Our simulation framework possesses a number of limitations. The first type of limitation pertains to the adopted negotiation setting. The studied negotiations have fewer offers, reflected in the low number of rounds. Moreover, we mainly looked at the final outcomes of the negotiations, while it is also possible to look at the gradual progression of the strategies leading to the final outcome. In this case, longer negotiation rounds are required as opposed to the average of 8 rounds for GPT-4 in our setting. Second, we only focus on one negotiation scenario. It is possible to use the proposed framework in other negotiation settings. In addition, the negotiation problem is relatively simple and could be rendered more complex by introducing additional variables, objectives, and constraints. Finally, the adopted economic models of the agents are simple (linear) and do not account for factors such as risk attitudes, time discounting, etc.

The second type of limitation is related to the way personality traits are defined. We uniformly sampled from the personality space to generate personality profiles (Eq. \ref{eq:personality_space}). One could claim that realistic personality traits are not uniformly distributed and depend on factors such as culture, geography, etc. Such specificities could also encode societal biases. Further research is needed to identify and mitigate potential biases in the simulated agents to ensure fairness across different demographic groups and their cultural and psychological specificities.

\section*{Ethics Statement}

This research on simulating negotiations using LLMs with synthesized personality traits raises several ethical considerations. Below, we list some of these considerations and propose some solutions.

\begin{itemize}

\item  While our study aims to understand negotiation dynamics with AI and develop beneficial negotiation assistants, the proposed techniques could be misused to manipulate or exploit people in real-world negotiations, especially in financial or business contexts, and potentially be adapted for malicious uses such as social engineering or fraud. We must ensure that this research is not used to create deceptive assistant AI agents. AI researchers must consider implementing safeguards against such misuse.

\item Although our study used synthetic data, future applications may involve training on accurate negotiation data from online marketplaces, which could raise serious privacy issues. Strict data protection and anonymization protocols should be implemented if real-world negotiation data is used.

\item The complexity of human psychology requires advanced statistical models to be validated experimentally before being implemented in AI systems.

\item As LLM agents become more sophisticated, we need to ensure transparency in their decision-making processes, especially in high-stakes negotiation scenarios. Methods for explaining the "reasoning" behind the agent's decisions should be developed.

\item The widespread adoption of negotiation agents could have significant economic effects, potentially altering market dynamics if deployed at a larger scale by many users. Such broader societal impacts should be carefully studied and mitigated.

\item Interacting with AI negotiators that closely mimic human personality traits can have unexpected psychological impacts on humans, particularly for vulnerable users (elders, children, minorities, etc.). Thorough user studies should be conducted to assess these effects.

\end{itemize}

To address such ethical concerns, we advocate interdisciplinary research between economists, psychologists, and AI experts to understand the long-term societal impacts of LLM-based negotiating agents. This should lead to ethical guidelines for developing such agents as well as the appropriate regulations governing their responsible use in banking and finance domains.

\section*{Acknowledgments}

This work was supported by JST ACT-X Grant Number JPMJAX23CP and JSPS Kakenhi Grant Number JP23K28145.

\bibliography{custom}

\appendix

\section*{Appendix}
\label{sec:appendix}

% \begin {enumerate}
%\item Personality IPIP tests. Figure \ref{fig:ipip}.
% \item Dialogue length (compare with CraigListBargain dataset)
%\item Strategy’s accumulation figure
%\item Llama results? (if not in 5.1 already)
%\item Offers data (PlotPrices/)
%\end {enumerate}

\section{Prompt Settings}
\label{PromptSettingsAp}

The section includes prompts given to the LLM agents, including the buyer agents, the seller agents, and the negotiation state detector. For each negotiation agent, we first specify the negotiation objectives and then the personality instructions through personality-describing adjectives. 
For the negotiation objectives, we provide the product name and the target price.
We also include a product description in the seller's instructions.

The instructions we used for the buyer are the following:

\begin{tcolorbox}[colback=gray!20!white, colframe=gray!80!black, title=\texttt{Prompt for buyer agent}] 
\vspace{-1mm}
\small
\texttt{Act as a buyer and try to strike a deal for a {[}PRODUCT{]} with a lower price through conversation. Your reply should not be too long. You would like to pay for {[}TARGET PRICE{]}. You can accept a higher price if the item is really good or there are other perks.\\
 \\
You have following personality: \\
{[}PERSONALITY DESCRIBING ADJECTIVES{]}\\
Reflect your personality in the negotiation process.
}
\end{tcolorbox}

%In addition to the product name and target price, we also include a product description in the seller's instructions. This product description is extracted from the CraigListBargain dataset. 
The instructions we use for the seller are the following:

\begin{tcolorbox}[colback=gray!10!white, colframe=gray!80!black, title=\texttt{Prompt for seller agent}] 
\vspace{-1mm}
\small
\texttt{Act as a seller that sells a {[}PRODUCT{]}, bargains with the buyer to get a higher deal price. Your reply should not be too long. our listing price for this item is {[}TARGET PRICE{]}. The detail of the product is the following: \\
{[}PRODUCT DESCRIPTION{]}\\
 \\
You have following personality: \\
{[}PERSONALITY DESCRIBING ADJECTIVES{]}\\
Reflect your personality in the negotiation process.
}
\end{tcolorbox}

From the dialogue context, the state detector extracts the negotiation state, the price offered by the current speaker, and the negotiation strategy of the current speaker (free text form). Note that the price is averaged if the participants are trying to strike a deal for more than two pieces of products. The prompt for the state detector is the following:

\begin{tcolorbox}[colback=gray!20!white, colframe=gray!80!black, title=\texttt{Prompt for state detection}] 
\small
\texttt{You will be given a partial dialogue in which a buyer and a seller negotiate about a deal.
Predict the average product price, dialogue state and the strategy of the {[}LAST SPEAKER ROLE{]} by the end of the dialogue.\\
 \\
{[}The dialogue{]}\\
seller: Hi, how can I help you?\\
 buyer: Hello, I'm interested in ...\\
...
}
\end{tcolorbox}

\section{Experiments with other LLM models}
\label{LLMApendix}
\begin{table*}[ht]
%\small
\centering
\resizebox{\textwidth}{!}{%
\begin{tabular}{l|cc|cc|cc|cc|cc}
\toprule
\textbf{Personality} & \multicolumn{2}{c|}{\textbf{EXT}} & \multicolumn{2}{c|}{\textbf{AGR}} & \multicolumn{2}{c|}{\textbf{CON}} & \multicolumn{2}{c|}{\textbf{NEU}} & \multicolumn{2}{c}{\textbf{OPE}} \\ \hline
  & \textbf{Buyer} & \textbf{Seller} & \textbf{Buyer} & \textbf{Seller} & \textbf{Buyer} & \textbf{Seller} & \textbf{Buyer} & \textbf{Seller} & \textbf{Buyer} & \textbf{Seller} \\ \hline
  
Intrinsic Utility & -0.095 & -0.068 & \negative{\textbf{-0.190**}} & -0.117 & 0.100 & 0.028 & 0.074 & -0.143 & -0.004 & \negative{\textbf{-0.220**}} \\

Joint Utility & -0.026 & 0.035 & 0.042 & -0.065 & -0.003 & 0.061 & -0.037 & -0.136 & \positive{\textbf{0.130**}} & -0.168 \\

Concession Rate & 0.042 & 0.100 & 0.064 & \positive{\textbf{0.215**}} & -0.006 & 0.146 & -0.006 & 0.113 & \positive{\textbf{0.096*}} & 0.076 \\

Success Rate & \positive{\textbf{0.208**}} & 0.155 & \positive{\textbf{0.263**}} & 0.132 & 0.039 & 0.114 & -0.023 & -0.063 & 0.054 & 0.131 \\

Negotiation Rounds & 0.055 & 0.097 & \negative{\textbf{-0.369**}} & \negative{\textbf{-0.188*}} & -0.034 & 0.004 & 0.088 & -0.059 & \positive{\textbf{0.143**}} & 0.109 \\
\bottomrule
\end{tabular}}
\caption{Spearman's correlation coefficients between negotiation metrics and Big Five personality traits when using GPT-3.5 as buyer and seller agents. Asterisks indicate statistical significance, with * denoting \(p < 0.1\) and ** denoting \(p < 0.05\).}
\label{table:gpt-3.5}
\end{table*}
\begin{table*}[ht]
%\small
\centering
\resizebox{\textwidth}{!}{%
\begin{tabular}{l|cc|cc|cc|cc|cc}
\toprule
\textbf{Personality} & \multicolumn{2}{c|}{\textbf{EXT}} & \multicolumn{2}{c|}{\textbf{AGR}} & \multicolumn{2}{c|}{\textbf{CON}} & \multicolumn{2}{c|}{\textbf{NEU}} & \multicolumn{2}{c}{\textbf{OPE}} \\ \hline
  & \textbf{Buyer} & \textbf{Seller} & \textbf{Buyer} & \textbf{Seller} & \textbf{Buyer} & \textbf{Seller} & \textbf{Buyer} & \textbf{Seller} & \textbf{Buyer} & \textbf{Seller} \\ \hline
  
Intrinsic Utility & -0.074 & -0.038 & \negative{\textbf{-0.236**}} & \negative{\textbf{-0.252**}} & 0.001 & 0.028 & \positive{\textbf{0.163**}} & -0.003 & \negative{\textbf{-0.111**}} & \negative{\textbf{-0.063*}} \\

Joint Utility & \positive{\textbf{0.089*}} & -0.115 & \positive{\textbf{0.151**}} & \negative{\textbf{-0.158**}} & -0.006 & 0.029 & \negative{\textbf{-0.178*}} & -0.003 & 0.076 & 0.050 \\

Concession Rate & \positive{\textbf{0.143**}} & \positive{\textbf{0.061*}} & \positive{\textbf{0.118**}} & \positive{\textbf{0.153**}} & \positive{\textbf{0.096*}} & 0.038 & \negative{\textbf{-0.212**}} & -0.025 & \positive{\textbf{0.113*}} & 0.039 \\

Success Rate & \positive{\textbf{0.089*}} & \positive{\textbf{0.120**}} & 0.055 & \positive{\textbf{0.121**}} & 0.024 & 0.043 & -0.087 &  -0.069 & 0.000& -0.047 \\

Negotiation Rounds & -0.065 & \positive{\textbf{0.096**}} & \negative{\textbf{-0.177**}} & \negative{\textbf{-0.151**}} & -0.045 & 0.040 & 0.024 &  -0.017 & 0.066 & -0.009 \\
\bottomrule
\end{tabular}}
\caption{Spearman's correlation coefficients between negotiation metrics and Big Five personality traits when using Llama-3 as buyer and seller agents. Asterisks indicate statistical significance, with * denoting \(p < 0.1\) and ** denoting \(p < 0.05\).}
\label{table:llama}
\end{table*}

While we mainly focused on the negotiation simulation based on the GPT-4 model, we also conducted experiments based on other types of LLMs:

\paragraph{GPT-3.5} We conducted the same experiments using GPT-3.5 (gpt-3.5-turbo-0125) as the buyer and seller agents. The Spearman's ranking correlation coefficients between personality traits and negotiation outcome are summarized in Table~\ref{table:gpt-3.5}. Compared to results based on GPT-4, we noticed that we cannot identify any significant correlations, indicating that GPT-3.5 is inferior in modeling personality-related behavioral patterns. 

\paragraph{Llama-3-70b} We also conduct the experiments using open-sourced Llama model (Meta-Llama-3-70B-Instruct). The Spearman's ranking correlation coefficients between personality traits and negotiation outcome are summarized in Table \ref{table:llama}. Compared to the results based on GPT-4 (Table~\ref{table:newCorr}), we found that many of the significant correlations have the same tendency. For instance, \textbf{AGR} shows a negative correlation with the intrinsic utility and negotiation length and a positive correlation with concession and success rate. 
% Llama-2-13b-chat-hf 
The above results show that our model can generalize across different LLMs.

\section{IPIP personality test}
\label{sec:ipip}

% Self-report personality tests
Self-report personality tests are widely used for assessing individual personality traits in psychological research. 
Previous works also used them as a direct measure of an LLM agent's personality \cite{pan2023llmspossesspersonalitymaking,serapio2023personality}.
Here, we conduct the personality test on the LLM agents we designed in Section~\ref{sec:agent_config} to evaluate whether they properly reflect the given synthetic personality profiles. 

% IPIP and prompt
We adopt the International Personality Item Pool (IPIP) 50 personality test~\cite{goldberg1992development}, which is a widely used personality inventory designed for assessing the Big Five personality traits.
The IPIP 50 test consists of 50 statements such as ``(I) am the life of the party'', ``(I) sympathize with others' feelings'', with each statement is related to one of the Big Five personality dimensions.\footnote{
The list of statements and scoring scheme we used in this work can be found at \url{https://ipip.ori.org/newBigFive5broadKey.htm}.
}
The test-taker responds to each of the statements using a 5-Likert scale ranging from ``1 = very inaccurate'' to ``5 = very accurate''.
We collect the response of the LLM agents to each statement with the following prompt:

\begin{tcolorbox}[colback=gray!20!white, colframe=gray!80!black, title=\texttt{Prompt for personality test}] 
\vspace{-1mm}
\small
\texttt{Act as person with following personality: \\{[}PERSONALITY DESCRIBING ADJECTIVES{]}\\
Evaluate the following statement:\\
{[}STATEMENT{]}.\\
\\
Please rate how accurately this describes you on a scale from 1 to 5 (where 1 = "very inaccurate", 2 = "moderately inaccurate", 3 = "neither accurate nor inaccurate", 4 = "moderately accurate", and 5 = "very accurate"). Please answer using EXACTLY one of the following:  1, 2, 3, 4, or 5.
}
\end{tcolorbox}

% scoring of IPIP.
Each Big Five personality dimension is measured by ten statements in the IPIP 50 test. To assess the tendency of a certain personality dimension, we average the scores of the corresponding statements. A higher average score indicates a stronger personality trait of this personality dimension.
%For instance, we can average an agent's responses for all the corresponding statements related to extraversion (EXT) and 
%We initialize Large Language Model (LLM) agents with randomly generated Big-Five personality profiles (Section ??\footnote{\ed{where is this?}}).
%While there are concerns about data contamination, we could use a self-reporting personality questionnaire to evaluate whether an LLM agent properly reflects its synthetic personality profile. 
We conduct the IPIP 50 test on $300$ agents with randomly assigned Big Five personality profiles and analyze the correlation between the given personality profile and the IPIP scores.

%\subsection{Personality Modeling Results}
% IPIP of GPT-4

\begin{figure}[t]
    % First Subfigure (Figure)
        \includegraphics[width=0.47\textwidth]{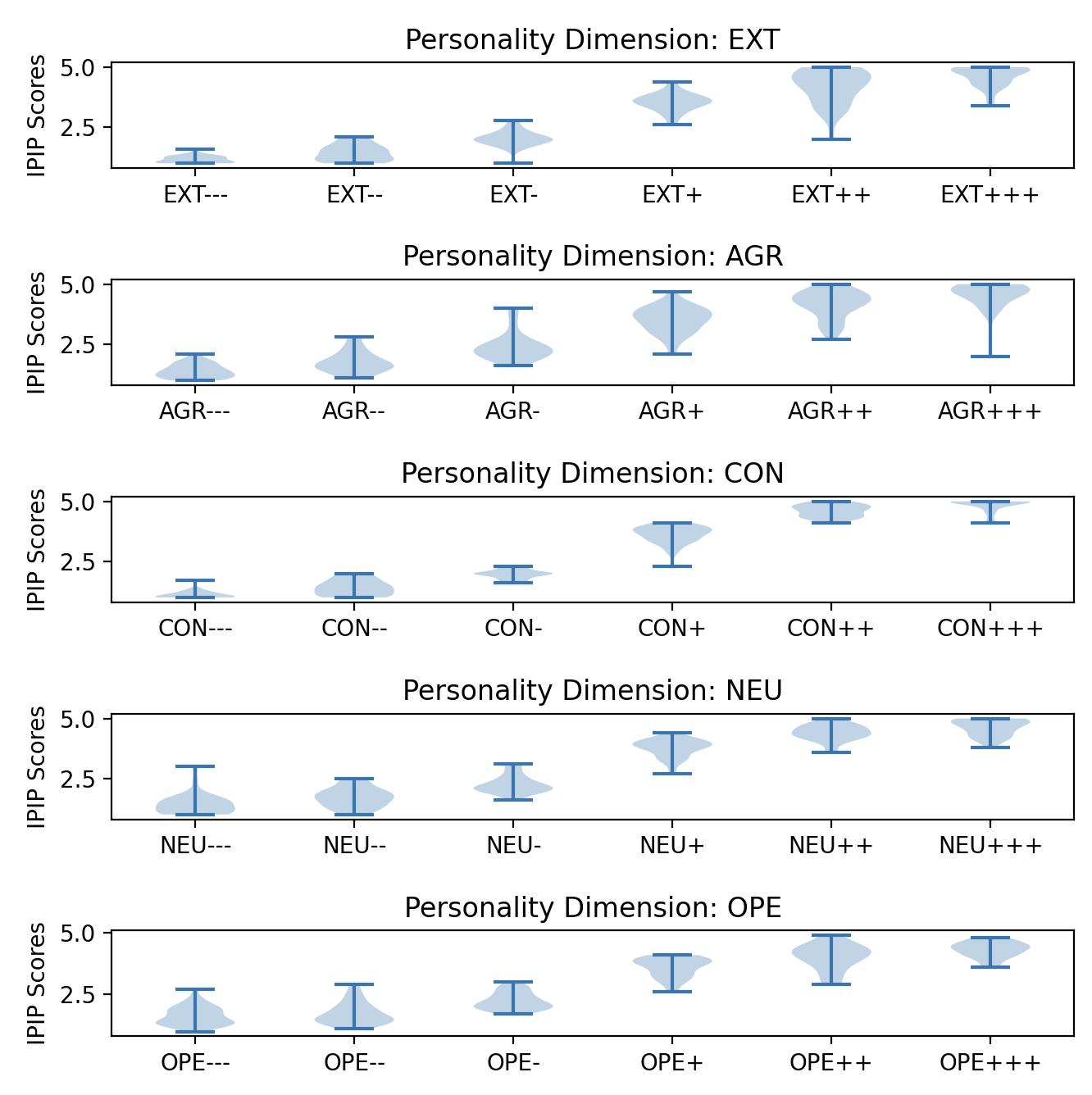}
        \caption{IPIP Personality test results of different personality levels.}
        %\vspace{0.5mm}
        \label{fig:ipip}
\end{figure}
\begin{table}[t]
    \centering
    \resizebox{\columnwidth}{!}{%
    \begin{tabular}{cccccc}
        \toprule
        & \textbf{OPE} & \textbf{CON} & \textbf{EXT} & \textbf{AGR} & \textbf{NEU} \\ \midrule
        IPIP OPE & \textbf{0.75*} & -0.02 & 0.04 & 0.09 & 0.04 \\
        %\hline
        IPIP CON & 0.02 & \textbf{0.74*} & \textbf{-0.20*} & 0.05 & -0.01 \\
        %\hline
        IPIP EXT & \textbf{0.14*} & \textbf{-0.16*} & \textbf{0.74*} & 0.11 & 0.01 \\
        %\hline
        IPIP AGR & -0.02 & \textbf{0.14*} & -0.07 & \textbf{0.73*} & 0.05 \\
        %\hline
        IPIP NEU & 0.06 & 0.03 & -0.10 & -0.10 & \textbf{0.78*} \\ 
        \bottomrule
    \end{tabular}
    }
    \caption{Spearman's rank correlation between synthetic personality scale and IPIP scores of the GPT-4 model. Asterisks indicate statistical significance, with * denoting p < 0.05.}
    \label{table:ipip}
\end{table}

Figure~\ref{fig:ipip} visualizes the average IPIP scores across agents with different polarities/degrees in this dimension in each Big Five dimension.
For all personality dimensions, we observe an increasing tendency of IPIP scores ranging from highly negative to highly positive personality profiles.
% correlation.
Table~\ref{table:ipip} shows the Spearman's correlation coefficients between the strength of the personality traits in the given personality profiles and the IPIP scores of each personality dimension (IPIP OPE, IPIP CON, IPIP EXT, IPIP AGR, and IPIP NEU). 
By looking at the diagonal elements, we can see a strong correlation between IPIP scores and the given personality profile in each dimension.
There are also some correlations across different personality dimensions.
For instance, assigning a higher openness personality to an LLM agent causes the extraversion scores (IPIP EXT) to rise.
However, these cross-dimension correlations are weaker compared to the diagonal ones.
We also include the results of Spearman's correlation test based on GPT-3.5 and llama-3-70b models in Table~\ref{table:ipip-gpt35} and Table~\ref{table:ipip-llama}, respectively.
The same tendency is observed across all three models.

The above evaluation verifies that the given personality profile is reflected by the LLM agents, at least in the context of a self-report personality test.
However, there remain concerns regarding data contamination affecting the validity of the evaluation based on these personality tests.
Since there are many publicly available datasets of the IPIP personality tests, it could be the case that the LLM is just memorizing these datasets as they are. If that is the case, when prompted with some statements in IPIP, the LLM will respond to the query by the memorization of the dataset. This would imply that the LLM only has a shallow understanding of the personality traits. %In this paper, we use the negotiation scenario to have an in-depth analysis of the personality simulation ability of LLMs.
On the other hand, our proposed method in this paper could serve as a complementary way to evaluate the LLM agents based on their behaviors.
For instance, all three models (GPT-4, GPT-3.5, and Llama-3-70b) can reflect the given personality traits in the context of the IPIP personality test (Table~\ref{table:ipip}, \ref{table:ipip-gpt35}, and \ref{table:ipip-llama}). However, only GPT-4 and Llama-3-70b models reflect the given personality traits in their negotiation behaviors (Table~\ref{table:newCorr} and \ref{table:llama}), but the GPT-3.5 models do not seem to reflect many significant behavioral patterns compared to the other two models (Table~\ref{table:gpt-3.5}).

% IPIP of GPT-3.5
\begin{table}[t]
    \centering
    \resizebox{\columnwidth}{!}{%
    \begin{tabular}{cccccc}
            \toprule
                & \textbf{OPE} & \textbf{CON} & \textbf{EXT} & \textbf{AGR} & \textbf{NEU} \\ 
            \midrule
            IPIP OPE & \textbf{0.59*} & 0.07 & \textbf{0.17*} & 0.04 & 0.03 \\
            %\hline
            IPIP CON & \textbf{0.16*} & \textbf{0.79*} & 0.00 & -0.11 & -0.04 \\
            %\hline
            IPIP EXT & 0.12 & -0.05 & \textbf{0.82*} & 0.12 & -0.05 \\
            %\hline
            IPIP AGR & 0.08 & -0.09 & \textbf{0.29*} & \textbf{0.72*} & -0.10 \\
            %\hline
            IPIP NEU & 0.05 & 0.02 & -0.10 & -0.05 & \textbf{0.84*} \\ 
            \bottomrule
    \end{tabular}
    }
    \caption{Spearman's rank correlation between synthetic personality scale and IPIP scores of the GPT-3.5 model. Asterisks indicate statistical significance, with * denoting p < 0.05.}
    \label{table:ipip-gpt35}
\end{table}

% IPIP of Llama-3-70b
\begin{table}[t]
    \centering
    \resizebox{\columnwidth}{!}{%
    \begin{tabular}{cccccc}
            \toprule
                & \textbf{OPE} & \textbf{CON} & \textbf{EXT} & \textbf{AGR} & \textbf{NEU} \\ 
            \midrule
            IPIP OPE & \textbf{0.88*} & 0.00 & 0.10 & -0.10 & 0.08 \\
            %\hline
            IPIP CON & -0.01 & \textbf{0.92*} & 0.01 & 0.05 & -0.09 \\
            %\hline
            IPIP EXT & 0.02 & \textbf{-0.18*} & \textbf{0.90*} & -0.13 & -0.10 \\
            %\hline
            IPIP AGR & 0.07 & 0.01 & \textbf{0.14*} & \textbf{0.85*} & -0.05 \\
            %\hline
            IPIP NEU & \textbf{0.14*} & -0.06 & \textbf{-0.16*} & 0.06 & \textbf{0.92*} \\ 
            \bottomrule
    \end{tabular}
    }
    \caption{Spearman's rank correlation between synthetic personality scale and IPIP scores of the Llama-3-70b model. Asterisks indicate statistical significance, with * denoting p < 0.05.}
    \vspace{-2mm}
    \label{table:ipip-llama}
\end{table}

%------------------------------------------------------------------------
% https://docs.scipy.org/doc/scipy/reference/generated/scipy.stats.chi2_contingency.html
% Seller: statistics=451.33, p_value=8.530553875069031e-53, dof=81
% Buyer: statistics=179.53, p_value=2.1724371086994425e-15, dof=54
% Joint: statistics=577.89, p_value=2.7313266465712027e-72, dof=90 

\end{document}